\documentclass[conference]{IEEEtran}
\IEEEoverridecommandlockouts
\usepackage{cite}
\usepackage{amsmath,amssymb,amsfonts}
\usepackage{algorithmic}
\usepackage{graphicx}
\usepackage{textcomp}
\usepackage{xcolor}
\usepackage{caption} 
\usepackage{subcaption} 
\usepackage{arydshln} 
\usepackage{booktabs} 
\usepackage{float} 
\usepackage[hidelinks]{hyperref}
\def\BibTeX{{\rm B\kern-.05em{\sc i\kern-.025em b}\kern-.08em
    T\kern-.1667em\lower.7ex\hbox{E}\kern-.125emX}}

\usepackage{fancyhdr}
\makeatletter
\let\old@headrule\headrule
\renewcommand{\headrule}{\if@fancyplain\let\headrulewidth\plainheadrulewidth\fi\old@headrule}
\makeatother

\renewcommand{\headrulewidth}{0pt}

\makeatletter
\def\ps@IEEEtitlepagestyle{%
  \def\@oddhead{\mbox{}\scriptsize\rightmark \hfil}%
  \def\@evenhead{\scriptsize\thepage \hfil \leftmark\mbox{}}%
  \def\@oddfoot{\hfil \mbox{}\parbox{5.5in}{\centering
  \footnotesize \textcopyright 2024 IEEE. Personal use of this material is permitted.
  Permission from IEEE must be obtained for all other uses, in any current or future
  media, including reprinting/republishing this material for advertising or promotional
  purposes, creating new collective works, for resale or redistribution to servers or
  lists, or reuse of any copyrighted component of this work in other works.}\hfil \mbox{}}%
  \def\@evenfoot{\mbox{}\parbox{5.5in}{\centering}\hfil \mbox{}\hfil}%
}
\makeatother
\IEEEoverridecommandlockouts

\begin{document}

\title{GEEvo: Game Economy Generation and Balancing with Evolutionary Algorithms
\thanks{This research was supported by the Volkswagen Foundation (Project: Consequences of Artificial Intelligence on Urban Societies, Grant 98555).}
}

\author{\IEEEauthorblockN{Florian Rupp}
\IEEEauthorblockA{\textit{Department of Computer Science} \\
\textit{University of Applied Sciences Mannheim}\\
Mannheim, Germany \\
f.rupp@hs-mannheim.de}
\and
\IEEEauthorblockN{Kai Eckert}
\IEEEauthorblockA{\textit{Department of Computer Science} \\
\textit{University of Applied Sciences Mannheim}\\
Mannheim, Germany \\
k.eckert@hs-manneim.de}
}

\maketitle

\begin{abstract}

Game economy design significantly shapes the player experience and progression speed. 
Modern game economies are becoming increasingly complex and can be very sensitive to even minor numerical adjustments, which may have an unexpected impact on the overall gaming experience.
Consequently, thorough manual testing and fine-tuning during development are essential. Unlike existing works that address algorithmic balancing for specific games or genres, this work adopts a more abstract approach, focusing on game balancing through its economy, detached from a specific game.

We propose GEEvo (Game Economy Evolution), a framework to generate graph-based game economies and balancing both, newly generated or existing economies.
GEEvo uses a two-step approach where evolutionary algorithms are used to first generate an economy and then balance it based on specified objectives, such as generated resources or damage dealt over time.
We define different objectives by differently parameterizing the fitness function using data from multiple simulation runs of the economy.
To support this, we define a lightweight and flexible game economy simulation framework.
Our method is tested and benchmarked with various balancing objectives on a generated dataset, and we conduct a case study evaluating damage balancing for two fictional economies of two popular game character classes.
\end{abstract}



\section{Introduction}


Games, whether analog or video games, that feel unbalanced are unsatisfactory to players, leading to boredom or frustration, and players will stop playing.~\cite{becker_what_2020}. A game's balance is thereby heavily influenced by the design and configuration of its internal economy which defines how virtual resources are created and can be transitioned to other resources.
A well-designed economy is a powerful system, incentivizing players to engage in certain behaviors to increase their virtual progress and to keep playing the game~\cite{schreiber_game_2021,adams_fundamentals_2014}.
This concept is present in many game genres such as in simulation games like \emph{Catan} or \emph{Factorio} (transitions of resources like wood or iron), first person shooters (health points and their relation to weapon damage), role playing games like \emph{The Legend of Zelda} or \emph{World of Warcraft} (e.g., mana and ability cool downs), or in general: the players' most valuable resource - their time.
Configuring the generation or requirement of resources for each transition significantly influences the entire system, making it challenging to achieve overall balance.
Even a small change can have large and sometimes unforeseen effects on the overall economic behavior and therefore on the entire playing experience. This is particularly the case when it comes to probabilistic mechanisms within the economic system, such as the type and amount of resource creation based on e.g., a dice roll. For these reasons, the design and balancing of game economies requires a lot of manual work, testing and fine-tuning in the development process~\cite{schreiber_game_2021}.



\begin{figure}
    \centering
    \includegraphics[width=1\linewidth]{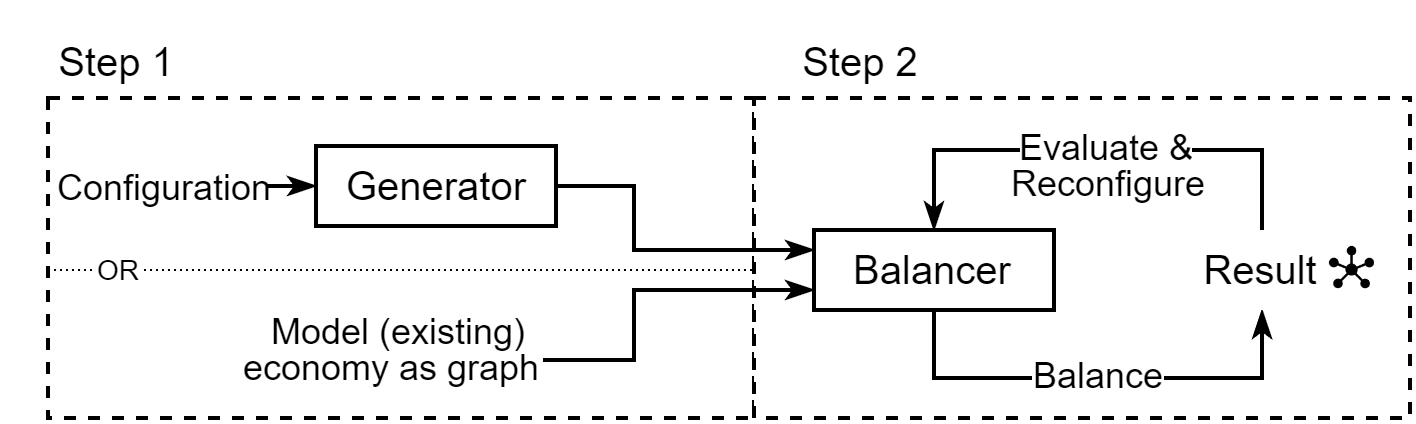}
  \caption{The process of GEEvo is two-step:  First, a game designer models an existing game economy using our simulation framework or creates one with the generator. Second, 
  the designer sets an objective based on which the balancer then optimizes the economy graph's weights. In an iterative process the designer evaluates the weights found and, if needed, may reconfigure the balancer. This may involve specifying static weights for e.g., a particular narrative context or enhancing the influence of probabilistic elements within the economy.
  }
  \label{fig:process}
\end{figure}

To support game designers in this important but time-consuming process, this work explores the algorithmic optimization of existing or generated game economies to meet various balancing objectives (e.g., character damage dealing).
Existing works focus on specific games or genres to address game balancing~\cite{volz_demonstrating_2016,sorochan_generating_2022,pfau_dungeons_2020,gravina_constrained_2016,mesentier_silva_evolving_2019}, whereas this research takes a different approach by modeling graph-based game economies at a more abstract level.

Therefore, we propose GEEvo (Game Economy Evolution, Figure~\ref{fig:process}), a framework to generate and balance graph-based game economies simulation-driven. 
By splitting up the generation and balancing into distinct units, we address the procedural content generation (PCG) and balancing problem separately.
Little work has been presented on the generation of game economies itself in literature. Rogers et al.~\cite{rogers_using_2023} introduce an evolutionary algorithm (EA) to generate graph-based game economies of different complexities for a simulation game. In this work, however, we address more complex economies including probabilistics for instance. 
Hence, we create the \emph{generator} — a controllable EA designed to generate random but valid economies. Simulations are conducted using a lightweight framework inspired by Machinations\footnote{https://machinations.io/}, which leverages the domain-specific language introduced in~\cite{klint_micro-machinations_2013}.
With this approach, an economy can be flexibly modeled as a graph with nodes representing functional components. 
The \emph{balancer}, a second EA, then optimizes the resource flow by adjusting edge weights in the economy graph.
With GEEvo, a designer e.g., has the capability to model, simulate, and balance the pace at which players progress towards a particular achievement, like crafting the mighty sword.
To assess our method, we evaluate the general performance on a set of generated economies and conduct a case study in balancing the damage output of fictitious economies of a mage and an archer. The Python code for GEEvo can be found on GitHub\footnote{https://github.com/FlorianRupp/GEEvo-game-economies}.
Our contributions are:
\begin{itemize}
    \item The \emph{generator}: An evolutionary algorithm for creating random but valid game economy graphs that are controllable in terms of output size and components.
    \item The \emph{balancer}: An evolutionary algorithm to balance graph-based game economies simulation-driven by optimizing its weights towards different objectives. 
    \item A lightweight and extendable framework to simulate game economies.
    \item A study to evaluate the balancing of generated economies with respect to different balancing objectives. In it, we demonstrate balancing using an example setting of damage dealing for a mage and an archer.
\end{itemize}

The paper is structured as follows: We give an overview of related work in Section~\ref{sec:related-word} and the method in Section~\ref{sec:method}. The latter includes the description of the economy simulation~framework~(\ref{sec:framework}), the generator~(\ref{sec:ea-gen}) and balancer~(\ref{sec:ea-balance}).
Experiments and results are described in Section~\ref{sec:experiments}, the discussion and limitations in~\ref{sec:discussion}, and the conclusion and future work in~\ref{sec:conclusion}.



\section{Related Work}
\label{sec:related-word}

Game economies have been studied in depth in the basic books~\cite{schreiber_game_2021} and~\cite{adams_fundamentals_2014}. A game's economy massively contributes to the overall game experience including its balancing and how fast players progress in the game. Especially through balancing the progress, free-to-play games use players' impatience for monetization~\cite{evans_economics_2016}.
Klint et al.~\cite{klint_micro-machinations_2013} introduce a domain specific language to flexibly model game economies as a graph which has been adapted further in~\cite{vanrozen_adapting}. The authors introduce different node types which create and transition resources. Based on the proposed design, we implement a more lightweight implementation to explore its compatibility with EAs in terms of generation and balancing.

In the area of games research, EAs are used extensively e.g., for game balancing~\cite{volz_demonstrating_2016,sorochan_generating_2022,gravina_constrained_2016,morosan_automated_2017}, playing a game~\cite{gaina_rolling_2017}, PCG tasks such as levels~\cite{togelius_search-based_2011,togelius_controllable_2013,shaker_evolving_2012}, scenarios~\cite{gerhold_computer_2023}, narratives~\cite{alvarez_tropetwist_2022}, rules~\cite{cook_mechanic_2013}, or textures~\cite{wiens_gentropy_2002} (cf. search-based PCG). A comprehensive and up-to-date survey about the usage of EAs for games is given by Togelius et al. in~\cite{togelius_evolutionary_2024}.

Whereas this work aims at balancing a game through its economy from a more abstract perspective, many works focus on a single game or genre to adjust game unit parameters or create new units using an EA. Volz et al. create balanced decks for a card game using an evolutionary multi-object optimization strategy~\cite{volz_demonstrating_2016}. To estimate a deck's win rate, the authors also use a simulation-driven approach, but supported by a statistical surrogate model. De Mesentier Silva et al. present an EA to balance card attributes for the game \emph{Hearthstone}~\cite{mesentier_silva_evolving_2019}.
Gravina et al.~\cite{gravina_constrained_2016} introduce the Constraint Surprise Search, a divergent evolutionary search method, to generate sets of diverse and balanced weapons for first person shooter (FPS) games.
Morosan et al.~\cite{morosan_automated_2017} use an EA to balance a real time strategy game (RTS) and Sorochan et al.~\cite{sorochan_generating_2022} to generate new and balanced units for an RTS game.
Another approach to balancing a game is through the procedural generation of the game map as shown for FPS games~\cite{lanzi_evolving_2014,lara-cabrera_balance_2014}. Automated game balancing, however, is not limited to EAs. Pfau et al. introduce data-driven deep player behavior models to replicate human behavior to adapt game parameters~\cite{pfau_dungeons_2020}. Furthermore, reinforcement learning is also popular e.g., for difficulty balancing~\cite{reis_automatic_2023} or the generation of balanced levels~\cite{rupp_balancing_2023}. We build on the idea presented in~\cite{rupp_balancing_2023} of splitting the generation and balancing step into two separate units and calculating the fitness in a simulation-driven manner.

Little work has been presented on the automated generation or balancing of game economies itself. Rogers et al. generate graph-based game economies using an EA~\cite{rogers_using_2023}. The authors thereby focus on the generation of economies with different perceived complexity levels and proof their results alongside with a user study. In their implementation the economy is represented by a tree structure where each resource is always forwarded to a resource conversion. In contrast to this work, we focus on the \emph{balancing} of the economy and allow for more complex economies including probabilities or supporting loops within in the graph.

Besides balancing, EAs in the context of games are applied for PCG tasks~\cite{shaker_procedural_2016} such as the generation of levels. A taxonomy and survey is given by Togelius et al.~\cite{togelius_search-based_2011}. In another work, Togelius et al. propose an EA to generate levels for an RTS game using a controllable multi-object approach~\cite{togelius_controllable_2013}.
Search-based methods like EAs, however, have a high dependency on randomness and are also computationally intensive at inference. Therefore, Khalifa et al. introduce mutation models that combine evolution with machine learning to enable faster generation of game levels~\cite{khalifa_mutation_2022}.
Taking an alternative approach, Khalifa et al. employ reinforcement learning for level generation, highlighting the advantage of fast inference once the model has been trained~\cite{khalifa_pcgrl_2020}.


\section{The GEEvo Framework}
\label{sec:method}

\begin{table*}
    \centering
    \caption{Overview of the different node types and their constraints to create valid game economies.}
    \begin{tabular}{lcccclll} \toprule 
         \textbf{Node types $T$} &  \textbf{Max in}&  \textbf{Max out}&  \textbf{Min out}&\textbf{Min in}&\textbf{Allowed inputs} & \textbf{Allowed outputs} & \textbf{Color} \\ 
        
        \cmidrule(lr){1-1} \cmidrule(lr){2-5} \cmidrule(lr){6-6} \cmidrule(lr){7-7} \cmidrule(lr){8-8}

         Source&  0&  3&    1&0&-&Pool, Random Gate &green\\  
         Random Gate& 1& 3&   2&1&Source, Converter&Pool, Converter&red\\ 
         Pool& 2& 3&   0&1&Source, Random Gate, Converter&Converter, Drain &blue\\ 
         Converter& 3& 1&   1&1&Pool, Random Gate&Pool, Random Gate&yellow\\ 
         Drain& 2& 0&   0&1&Pool&- &orange\\ \bottomrule
    \end{tabular}
    \vspace{-1mm}
    \label{tab:node-types}
\end{table*}

The two-fold process of GEEvo is shown in Figure~\ref{fig:process}. We use two EAs, one for the constraint-based generation of game economies and a second one for the balancing of an economy's weights according to a given balancing objective. As applied in~\cite{rupp_balancing_2023}, we also separate the generation and balancing processes into two separated units for better performances of both.
Additionally, the encapsulation allows the application of the balancer to already existing economies.
The generator is controllable in terms of the economy's graph's size and the distinct number of each node type. The balancer optimizes an economy's weights toward a balancing objective by simulating the economy to calculate the fitness of its current weights.

Before we describe both algorithms in detail in Sections~\ref{sec:ea-gen} and~\ref{sec:ea-balance} including an overview of the different fitness functions, we present our economy simulation framework and its application.

\subsection{Game Economy Simulation Framework}
\label{sec:framework}

\begin{figure*}
\centering
  \begin{subfigure}[b]{0.45\textwidth}
    \centering
    \includegraphics[width=\textwidth]{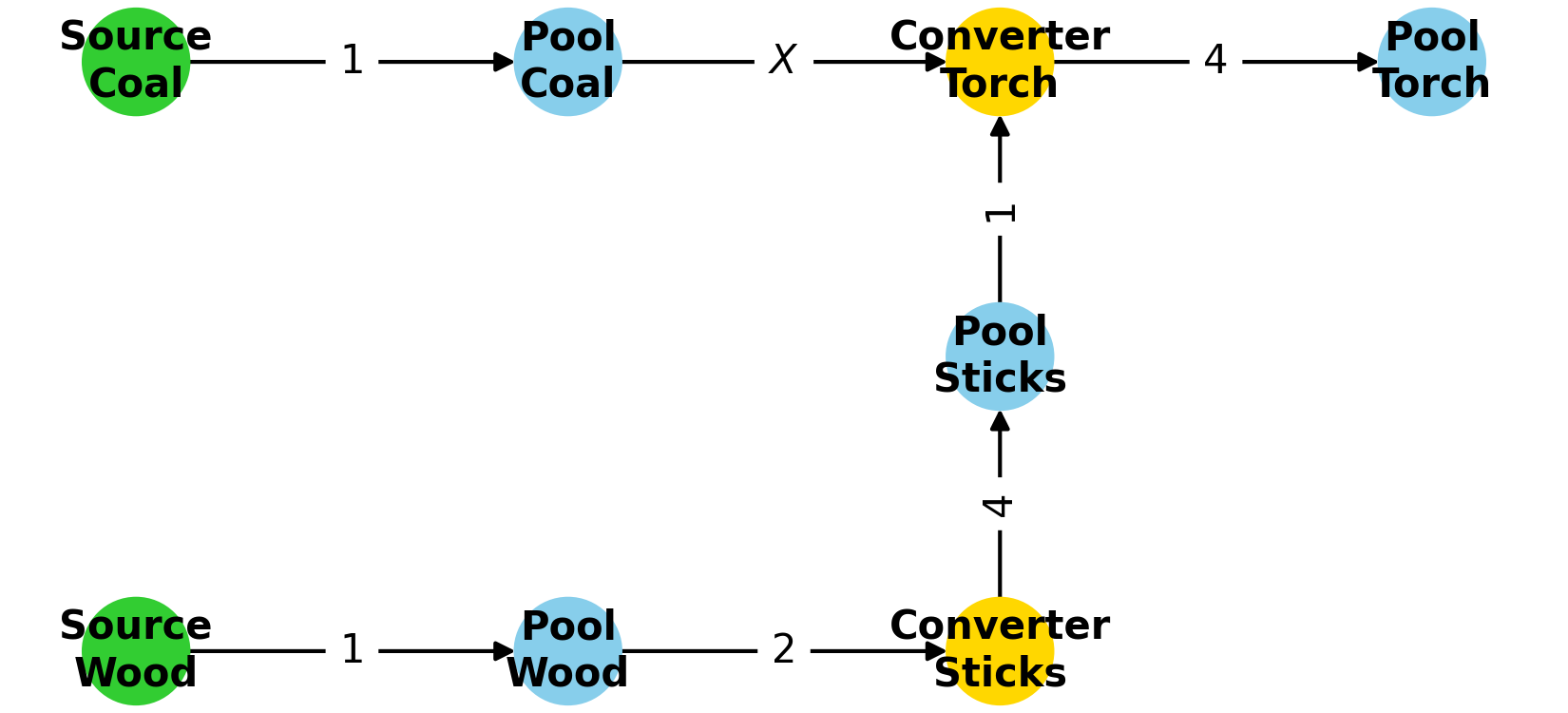}
    \caption{Example economy from the game Minecraft for crafting torches from basic wood and coal resources.}
    \label{fig:example-economy}
    \vspace{0.8cm}
  \end{subfigure} 
  \begin{subfigure}[b]{0.5\textwidth}
  \centering
    \begin{subfigure}{0.95\textwidth}
        \centering
      \includegraphics[width=\textwidth]{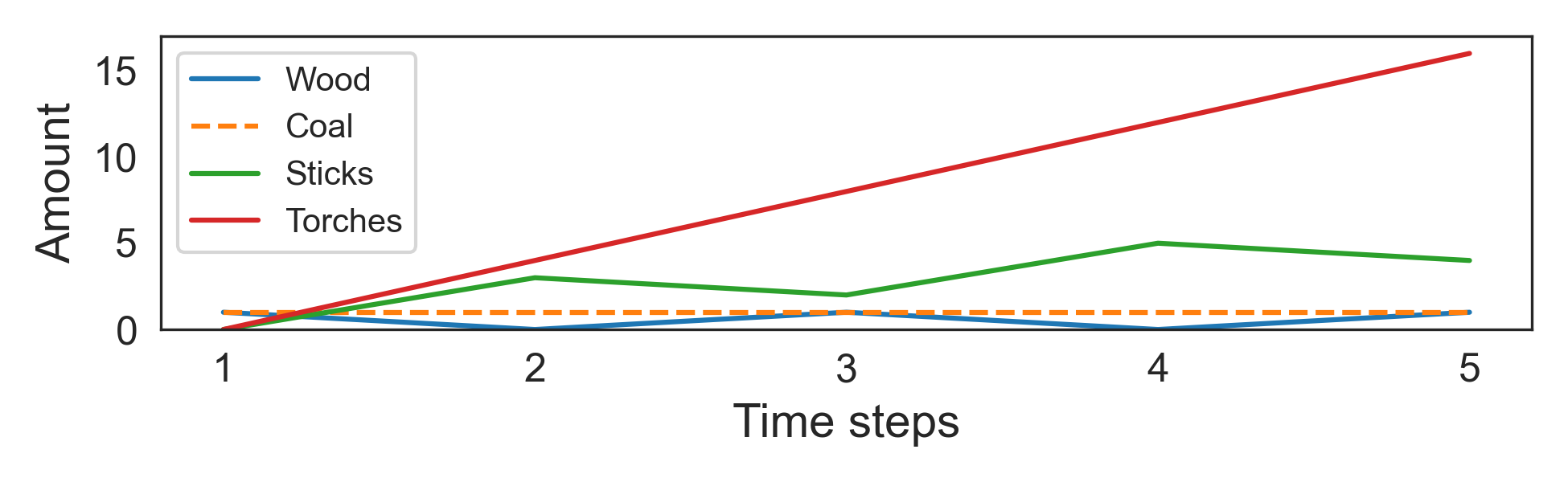}
    \vspace{-6.5mm}
      \caption{Simulation 1: $X=1$ (original value).}
      \label{fig:example-economy-sim1}
    \end{subfigure}
    \begin{subfigure}{0.95\textwidth}
    \centering
      \includegraphics[width=\textwidth]{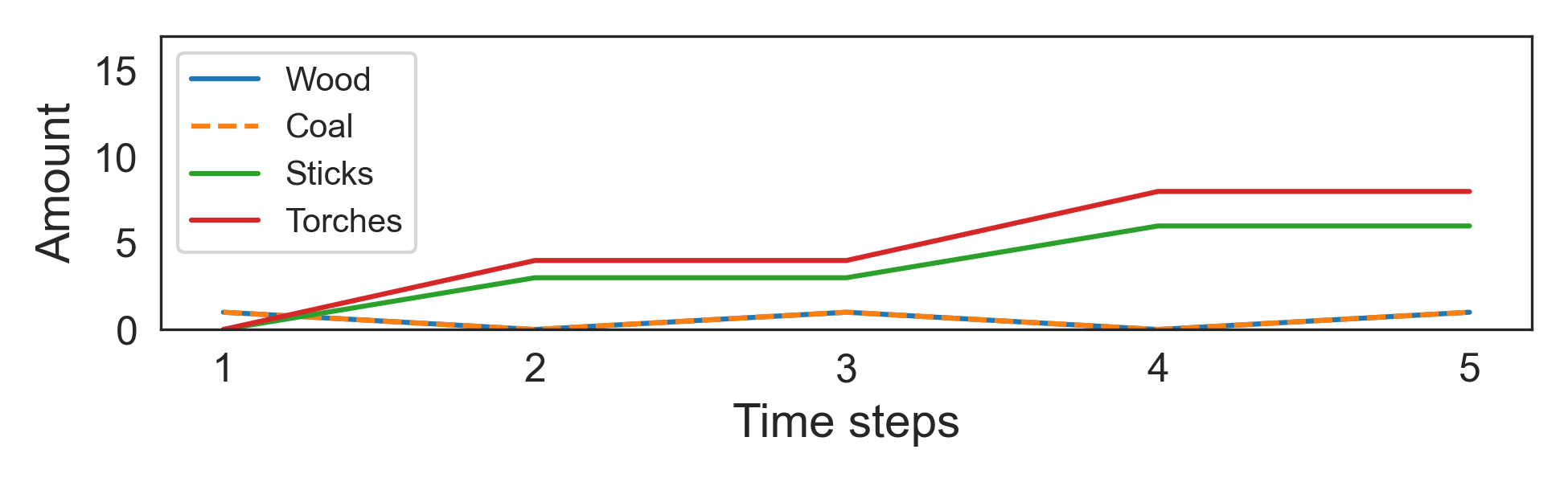}
    \vspace{-6.5mm}
      \caption{Simulation 2: $X=2$ (value for demonstration).}
      \label{fig:example-economy-sim2}
    \end{subfigure}
  \end{subfigure}
  \caption{Example of a game economy using the proposed framework and two simulations of it, each with a different configuration. The graph (a) shows the economy from the game Minecraft to craft torches from wood and coal in an automation setting. (b) and (c) show the monitoring of the pool nodes simulating the economy in (a). By only changing the amount of coal needed to craft torches, the entire economy behaves differently.}
\end{figure*}

A game economy can be considered as a directed graph which defines how resources of different types are generated and transitioned to other resource types. We only deal with fixed economic systems, where all resources come from the economy itself.
In this context, nodes in that graph represent different functional components such as the creation of resources or conversion to different types. Weights on the edges describe how many resources flow from one node to the adjacent one. Depending on the node type, the weights are absolute values or probabilities.

Our framework defines five different node types. To create valid and executable economy graphs, we establish constraints for each node type, specifying the permissible connections to other node types (Table~\ref{tab:node-types}). The values for maximum (max) and minimum (min) output sizes are each chosen to keep the economies more manageable. However, these values could also be changed to allow for more complex economies (e.g., increase max output).

\paragraph{Source} Sources are entry points creating resources and adding them to the economy.
\paragraph{Random gate} A random gate distributes incoming resources based on the probabilistic weights of its outgoing edges. It must be ensured that the sum of the weights of all outgoing edges equals one. This can be used to model e.g., critical attacks or random drops.
\paragraph{Pools and fixed Pools} Pool nodes have an intern memory to store incoming resources. They serve as buffers for outputs from sources, random gates, or converters or as end points of the economy. For the analysis of the economy, we can monitor the fluctuations of the resources within pools over time. Fixed pools, a subform of a pool buffer a maximum of the number of resources equal to the highest outgoing weight. This is particularly useful when modeling ability cooldowns.
\paragraph{Converter} A converter transitions one or multiple incoming resources to one outgoing resource.
\paragraph{Drain} Drains permanently remove resources from the economy. As with pools, drains can be monitored.

An example of how to model an existing economy from the sandbox game Minecraft~\cite{mojang_mincraft_2011} is given in Figure~\ref{fig:example-economy}. It shows the torch crafting process in an automation setting where specific amounts of resources are added to the economy via sources per time step. Using pool and converter nodes we can define this economy for torches from wood and coal sources. While coal can be directly used for crafting torches, the wood resource must first be converted into sticks. According to the original implementation\footnote{https://www.minecraft-crafting.net/}, the conversion to sticks yields four sticks per two wood entities. In this example, other resources such as the time to collect resources or the need for a crafting table for resource conversion are neglected.

\emph{Execution of Game Economy Simulations:}
The directed graph $G$ of a game economy can be denoted as $G=(V,E)$, where $V$ are its vertices (nodes) and $E$ are the edges connecting vertices. For each vertex $v_i \in V$ a subset $A^{v_i} \subset E$ exists containing all outgoing edges. The types of all allowed node types (such as source or pool) are described in the set $T$ (cf. Table~\ref{tab:node-types}).
The execution of the graph economy is then done recursively. For each $v_i$ all edges $e_j \in A^{v_i}$ are executed by a function $h_{\tau}(v_i, e_j),\, \tau \in T$
, $h_{\tau}$ respectively to the type $\tau$ of $v_i$.

We plot two example courses of simulations with different weights of the economy in Figure~\ref{fig:example-economy} in the Figures~\ref{fig:example-economy-sim1} and~\ref{fig:example-economy-sim2}. For this example, we assume both resources, wood and coal, to create one of each per time step. By changing the needed amount $X$ for coal to converse coal and sticks to torches, this example demonstrates how small changes to a single weight can impact the whole economy. While e.g., for $X=1$ the number of torches grows linearly per time step; for $X=2$ it is gradual. Also, the curve progressions of available wood and coal is different.

\subsection{Evolutionary Generation of Game Economies}
\label{sec:ea-gen}
The generator creates valid game economy graphs within the framework outlined in Section~\ref{sec:framework}. A valid economy graph must be weakly connected and adhere to the constraints in Table~\ref{tab:node-types}. The generator's task thereby is to connect nodes with edges, meeting all constraints. It operates by defining a population of individuals and iteratively optimizing them through mutations over multiple generations. The generator is designed for controllability, allowing users to specify the number and types of vertices in the generated graph. For instance, it can generate an economy with three sources, two random gates, one converter, and four pools.
The execution of the algorithm stops if a valid graph has been found or a maximum of allowed steps is exceeded.

\subsubsection{Initialization and Population} During initialization, all vertices defined in the external configuration are initialized depending on their type. The population consists of a configurable number of individuals, with a single individual representing the edge list of the graph. After initialization, the edge lists of all individuals are empty and will be filled iteratively in the execution.
\subsubsection{Mutations} Mutations are the driving force to evolve the graph.
Since the performance is satisfying (cf. Section~\ref{sec:ea-gen}), we abstain the implementation of a crossover. In each generation, for each individual, two vertices are randomly selected to add an edge. If this edge is allowed according to the constraints, it is added to the individual, otherwise it is not. This simple and greedy approach may get stuck, since at some point the graph is still not valid and no valid edges are allowed anymore. To address this shortcoming, a second mutation, may in each generation, remove a previously created edge from a random individual with a certain probability.
\subsubsection{Fitness function} The fitness function (Eq.~\ref{eq:fitness}) embeds the constraints to ensure the validity of the graph $G$ with its vertices $V$. It creates the graph based on the created edges and sums up the number of dissatisfied constraints for each vertex $v \in V$. This is done using the auxiliary function $validate(v)$. The function is defined in the interval $[0,\infty)$, where 0 represents the maximum fitness of an individual; to achieve best fitness the function must therefore be minimized.

\begin{equation}
\label{eq:fitness}
fitness(G) = \sum_{v\,\in\,V}{\, validate(v) }
\end{equation}

\subsection{Evolutionary Balancing of Game Economies}
\label{sec:ea-balance}

\begin{figure}
    \centering
    \includegraphics[width=0.95\linewidth]{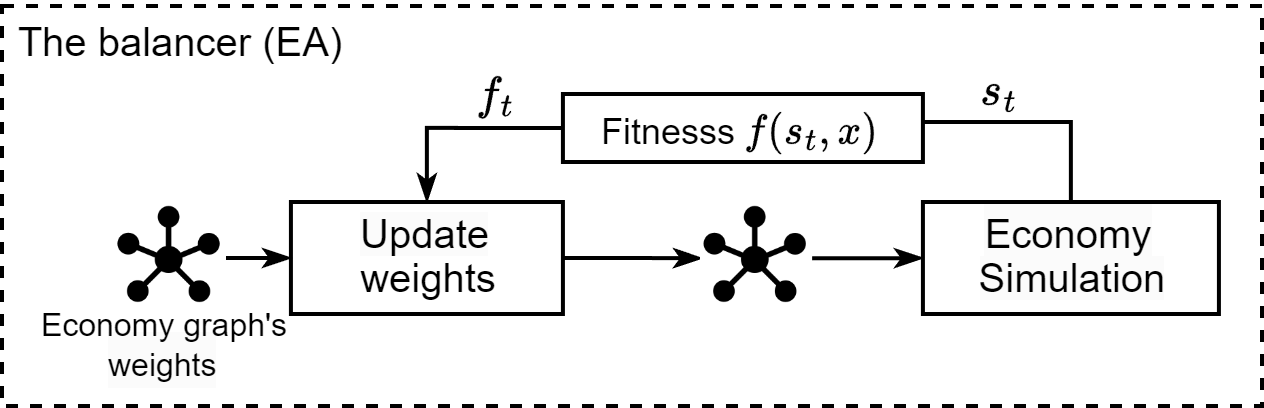}
    \caption{The structure of the balancer in detail: The balancer iteratively optimizes an economy's weights toward a balancing objective $x$. Therefore, it adjusts the weights based on the fitness $f_t$ per time step $t$ in relation to $x$ through crossovers and mutations. $f_t$ is calculated based on the result $s_t$ of multiple simulation runs of the economy.}
    \label{fig:architecture-balancer}
\end{figure}

The balancer optimizes the weights of one or multiple economies towards a balancing objective which is expressed by parameterizing its fitness function. An overview of the balancer's internal structure is given in Figure~\ref{fig:architecture-balancer}.
Utilizing crossovers and mutations, the balancer optimizes its population, consisting of multiple individuals. Following crossover and mutations, the population undergoes sorting based on fitness, retaining only individuals up to the population size for the subsequent generation.
The algorithm terminates either when the fitness is at its best for a single individual, or the maximum number of generations is exceeded.
In some narrative settings, a designer wants to keep specific values. Therefore, the balancer can be set up with static weights that remain unalterable throughout the balancing process. An example on the application is given in the case in Section~\ref{sec:case-study}.

\subsubsection{Individuals and population} A population consists of $n$ individuals where each represents all weights of an economy. It can be thought of as a list where each weight's index is mapped to the same index of the graph's edge list. In case of balancing two economies at once, an individual consists of the two weights lists of the respective economies. At the initialization of an individual, its weights are set randomly where all values must be greater than 0. 

\subsubsection{Crossover}
In each generation all individuals of the current population are paired randomly for crossover. For each pair, we iterate over both individuals' $k$ and $l$ weights $w_{ki}, \, w_{li}$ simultaneously and apply one of four randomly selected operations, each with equal probability. The four operations for the new weight at index $i$ are choose $w_{ki}$ or $w_{li}$ directly, $w_{ki} + w_{li}$ or, $w_{ki} - w_{li}$. After crossover, two individuals each produce a new child individual.


\subsubsection{Mutations}
For each mutation a random individual is selected from the population and a random weight is chosen. This weight is then modified by either adding or subtracting the random number by 0.5 each. If subtracting would yield a value $<0$ it is set to 1. 

\subsubsection{Fitness functions}
\label{sec:fitness-functions}
With the fitness function we define the goal towards which an economy should be balanced. All fitness is computed by observing the state $s_{t}^{p_j}$ (the amount of contained resources) of one or multiple pool nodes $p_j \in P, P \subset V$ at a selected time step $t, t \leq n$ within the $[1,n]$ steps of a simulated economy. 
To mitigate randomness, we run each simulation with the same weights $m$-times (see Section~\ref{sec:m-sim}).
\paragraph{Balancing a resource to an absolute value} This function (Eq.~\ref{eq:f1}) is motivated by adapting the weights of an economy so that a selected $p_j$ equals a given value $x$ after a fixed number of time steps $n$. One possible use case is to balance the economy to be capable of producing a distinct amount ($x$) of a resource ($p_j$) within a given time period ($n$). Such a resource can e.g., be coal or damage points, depending on the setting.
Since $s_{t}^{p_j}$ is based on stochastic simulations, we add an additional parameter $\alpha$ as a threshold value to the average so that the algorithm can also terminate at values close to the maximum fitness. As we will later explore (cf. Section~\ref{sec:case-study}), the configuration of this parameter is further important to control the influence of randomness in an economy when balancing.
Eq.~\ref{eq:f1} is defined in the interval $[0,1+\alpha]$, where $1+\alpha$ represents the maximum fitness of an individual. The auxiliary function $prop$ (Eq.~\ref{eq:f}) computes the proportion of $x$ and $s_{t}^{p_j}$.

\begin{equation}
\label{eq:f1}
f_1(s_{t}^{p_j}, x) = 
    \alpha + \frac{1}{m} \sum\limits_{i=1}^{m} prop(s_{ti}^{p_j},x), \quad p_j \in P
\end{equation}

\begin{equation}
\label{eq:f}
prop(s_{t}^{p_j}, x) = 
\begin{cases} 
     s_{t}^{p_j} \cdot \cfrac{1}{x},\quad x > 0 & \text{if } x > s_{t}^{p_j} \\[9pt]
     x \cdot \cfrac{1}{s_{t}^{p_j}},\quad s_{t}^{p_j} > 0 & \text{if } x \leq s_{t}^{p_j}
\end{cases}
\end{equation}


To balance two resources $s_{t}^{p_j}$ and $s_{t}^{p_k}$ within the same economy instead of a single absolute value we can use Eq.~\ref{eq:f1}, but with a different parametrization: $f_1(s_{t}^{p_j}, s_{t}^{p_k}), p_j,p_k \in P$.



\paragraph{Balancing two resources of different economies to the same value} Function $f_2$ parameterizes Eq.~\ref{eq:f1} to balance two resources of two different economies $\theta$ and $\phi$. We apply this function in the case study in Section~\ref{sec:case-study} to balance the dealt damage of a mage and an archer class within the same time period:

\begin{equation}
\label{eq:f3}
f_2(s_{t}^{p_j}, s_{t}^{p_k}) = f_1(s_{t}^{p_j}, s_{t}^{p_k}), \quad p_j \in P_\theta, \;p_k \in P_\phi
\end{equation}


\label{sec:m-sim}

\section{Experiments and Results}
\label{sec:experiments}
The evaluation is twofold: First, we investigate on the general performance of the generator and second, on the balancer including a case study.

\subsection{Evaluation of Game Economy Generation}
To evaluate the generator, we create a set of 200 economy graphs with a number of nodes in the range of 5 to 20. Also, the distribution of different node types is randomly varied. Within 50k iterations, the algorithm could generate valid graphs according to the constraints and the external configuration in 97\% of cases. The generated dataset therefore consists of 194 graphs.
The median number of iterations required to complete is 641 (90\% quantile: 9354), the median running time is 25~ms (90\% quantile: 296 ms)\footnote{We use for all experiments an AMD EPYC family 23 model 1 processor with 2.6 GHz.
One core is assigned per execution of a single graph.}. 

We use this set of economy graphs for the evaluation of the balancer with different balancing objectives.

\subsection{Evaluation of Game Economy Balancing}

\begin{figure*}
\centering
  \begin{subfigure}[b]{0.45\textwidth}
    \centering
    \includegraphics[width=\textwidth]{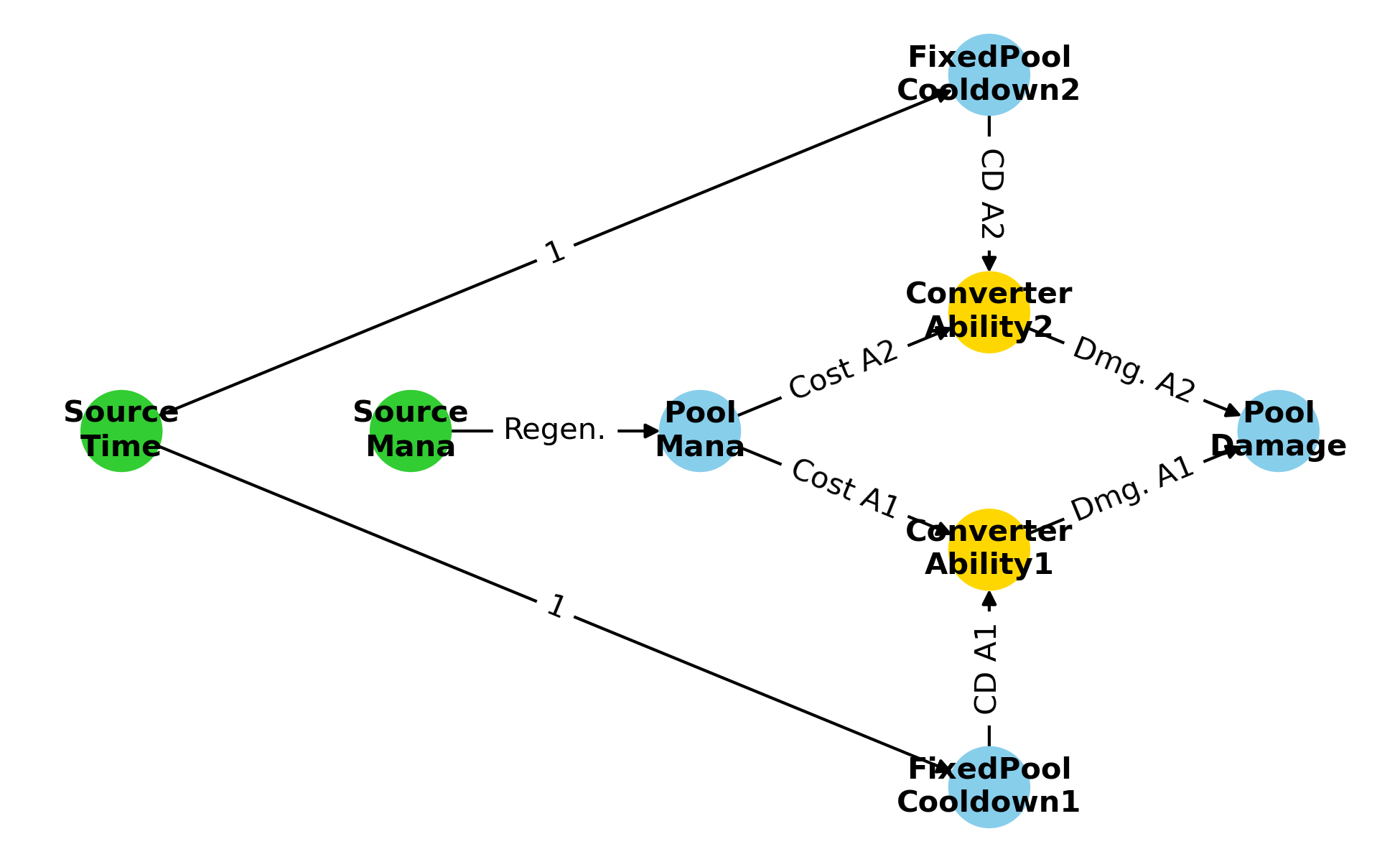}
    \caption{The mage's economy graph. Seven values can be balanced accordingly: the cooldowns (CD), mana costs (Cost A1, A2), and damage values (Dmg.) for both abilities (A1 and A2); as well as the overall amount of mana regenerated (Regen.) per time step.}
    \label{fig:economy-mage}
  \end{subfigure}
  \hspace{0.4cm}
  \begin{subfigure}[b]{0.45\textwidth}
      \includegraphics[width=\textwidth]{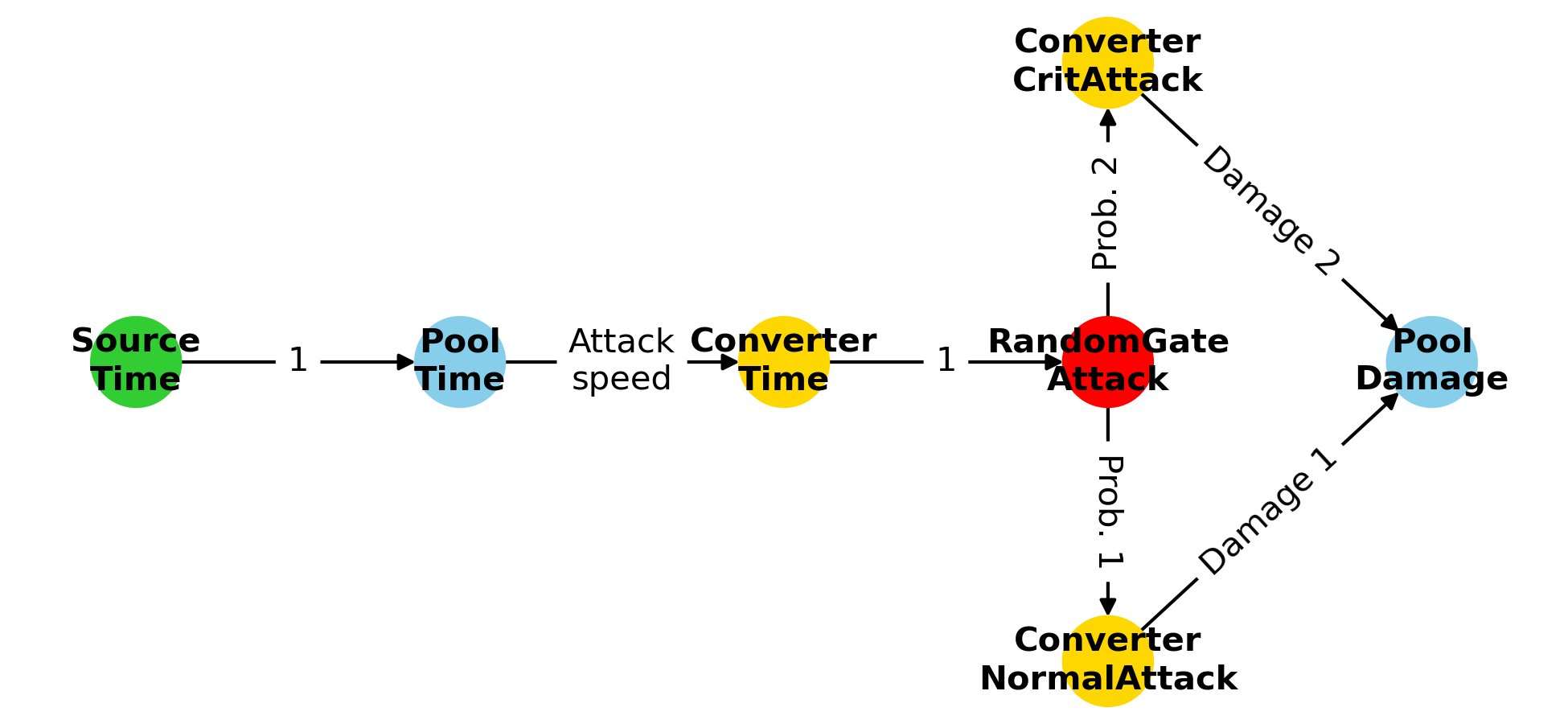}
      \hspace{2cm}
      \caption{The archer's economy graph. Five values can be balanced accordingly: the overall attack speed, the probabilities (Prob.) to perform a normal or critical attack and the the damage values for each.\\}
      \label{fig:economy-adc}
    \end{subfigure}
  \caption{Economy graphs for the case study to balance the damage dealing of a mage (a) and an archer (b). The values to be balanced are the weights on the edges. Fixed values are represented by absolute values.}
  \label{fig:economies-case-study}
  \vspace{-4mm}
\end{figure*}

\subsubsection{General Performance with different balancing objectives}

To assess the balancer's overall performance, we apply it to the previously generated dataset, focusing on balancing a randomly selected pool of an economy to a specific value after a defined number of simulation steps. To test the algorithm's adaptability to varied values, both the number of simulation steps and the specific target value are randomly chosen within the intervals $[10,30]$ and $[20,100]$, respectively. An example would be balancing the pool for torches in Figure~\ref{fig:example-economy} to a value of 28 after 16 simulation steps.
Therefore, we employ the fitness function (Eq.~\ref{eq:f1}) introduced earlier.
Given the probabilistic nature of random gates, achieving maximum fitness is often impossible; hence, we compare the results for different values of $\alpha$. The timeout for the algorithm to stop is set to 500 iterations. The size of the population is set to 20 and we run ten simulations with the same weights each generation for each individual. This results in 300 simulation runs per generation alone. The computational effort involved in carrying out the simulation is acceptable. Conducting ten simulations per economy and generation consistently results in a computing time of less than 284 ms.

Table~\ref{tab:gen-performance} presents the results, indicating a significant variation in the proportion of balanced economies based on~$\alpha$. As the threshold $\alpha$ increases, so does the proportion of balanced economies, including those that were initially balanced. Additionally, the median execution times and generations are dependent on $\alpha$, with higher values of $\alpha$ leading to solutions being found in fewer generations, thereby reducing overall computation time.



\begin{table}
    \centering
    \caption{Results for balancing the generated dataset of economy graphs using fitness function 1 (Eq.~\ref{eq:f1}) towards different values of $\alpha$.}
    \begin{tabular}{lccc}
    \toprule
            &  $\alpha=0.05$ &  $\alpha=0.01$ & $\alpha=0.0$ \\ \cmidrule(lr){1-1} \cmidrule(lr){2-2} \cmidrule(lr){3-3} \cmidrule(lr){4-4}
         \textbf{Balanced (\%)} &  93.3&  83& 58.8\\
         \textbf{Improved (\%)} & 77.3& 88.7&94.8\\
         \textbf{Initial balanced (\%)} &  27.3&  8.8& 2.5\\
         \textbf{Median generations} &  1&  7& 196\\
         \textbf{Median execution time (s)} &  18,4&  66& 703.2\\
         \bottomrule
    \end{tabular}
    \label{tab:gen-performance}
    \vspace{-2mm}
\end{table}

\subsubsection{Case study: Balancing a Mage with an Archer}
\label{sec:case-study}

Many game genres (e.g., MOBA: Multiplayer Online Battle Arena) offer players the opportunity to choose from different characters, each assigned to specific classes.
While each character has a unique game design, characters of the same class have a similar play style. The different character designs offer players various strategies to win the game. However, in order for a game to be balanced, it is necessary to ensure that different strategies, if played well, are viable to win the game~\cite{schreiber_game_2021}.

In this case study, we examine GEEvo for its ability to balance two different economies, using two popular classes as an example: a mage and an archer. For each class, our goal is to achieve a comparable maximum damage output within a specified time frame. Therefore, we use Eq.~\ref{eq:f3}.
Both economy graphs are shown in Figure~\ref{fig:economies-case-study}.
The mage's game design is based on casting two different abilities (Fig.~\ref{fig:economy-mage}). Each ability has a specific mana cost, a cooldown, and a value for the damage it deals. Mana is generated at each time step. So, there are seven values which need to be balanced accordingly: the cooldowns, mana costs, and damage values for both abilities; as well as the overall amount of mana regenerated per time step.
In contrast, the archer is designed to perform attacks that have a cooldown based on its attack speed (Fig.~\ref{fig:economy-adc}). Each attack has a chance to deal additional damage (cf. critical damage). In this economy five values can be adjusted for balancing: the probabilities for a normal attack and critical damage, the damage values for both, and the attack speed. Given the narrative context during the cooldown modeling for both economies, we establish these weights as static, allowing us to manually set the value to one (cf. Section~\ref{sec:ea-balance}).

We experiment with two different values for the fitness threshold $\alpha$, each time with the same seed. To mitigate randomness, we run ten simulations with a length of 30 time steps per generation and a population size of ten.
The algorithm terminates after two generations with a total computation time of 1.4 seconds for $\alpha=0.05$. For $\alpha=0.01$ the algorithm terminates after six generations in 16.6 seconds.

\begin{table}[]
\centering
\caption{Results after balancing attributes for the mage and archer economy compared for two different values of $\alpha$ with the goal to deal the same damage within a given period of time.}
\begin{tabular}{l@{\hspace{2.3mm}}ll@{\hspace{2.3mm}}ll@{\hspace{2.3mm}}ll@{\hspace{2.3mm}}l}
\toprule
\multicolumn{4}{c}{Result $\alpha$=0.05}  & \multicolumn{4}{c}{Result $\alpha$=0.01} \\ \cmidrule(lr){0-3} \cmidrule(r){5-8}
\textbf{Mage}          & $w$  & \textbf{Archer} & $w$  & \textbf{Mage}           & $w$   & \textbf{Archer}       &  $w$ \\ \cmidrule(lr){0-1} \cmidrule(lr){3-4} \cmidrule(lr){5-6} \cmidrule(lr){7-8}
M. Reg.    & 3     & A-Speed     & 2    & M. Reg.     & 2       & A-Speed     & 1 \\
A1 CD      & 1     & A1 Prob.    & 0.88 & A1 CD       & 1       & A1 Prob.    & 0.76 \\
A1 M.      & 3     & A1 Dmg.     & 1    & A1 M.       & 3       & A1 Dmg.     & 2 \\ 
A1 Dmg.    & 3     & A2 Prob.    & 0.12 & A1 Dmg.     & 3       & A2 Prob.    & 0.24 \\
A2 CD      & 3     & A2 Dmg.     & 3    & A2 CD       & 2       & A2 Dmg.     & 2 \\
A2 M.      & 2     &             &      & A2 M.       & 2       &             &   \\
A2 Dmg.    & 3     &             &      & A2 Dmg.     & 3       &             &   \\ \cmidrule(lr){0-3} \cmidrule(r){5-8} \morecmidrules \cmidrule(lr){0-3} \cmidrule(r){5-8}  
\addlinespace[2pt]
\textbf{$\sum$ Dmg.} & 55    &   \multicolumn{2}{c}{53.8$\pm$4.8}      & \multicolumn{2}{c}{60} &   \multicolumn{2}{c}{60$\pm$0} \\ \cmidrule(lr){0-3} \cmidrule(r){5-8}

\multicolumn{1}{l}{\textbf{Fitness}} &  \multicolumn{2}{c}{0.95 + $\alpha$} &   &   \multicolumn{4}{c}{1.0 + $\alpha$} \\
\bottomrule
\vspace{-5.5mm}
\label{tab:mage-archer}
\end{tabular}
\end{table}

Results of the found attributes (weights) are displayed in Table~\ref{tab:mage-archer}.
For both runs, the balancer finds a solution and terminates within the permitted number of generations. The weights found for the mage are both comparable, whereas those for the archer differ mainly for the probabilistic values for a critical hit. With $\alpha=0.05$, a critical hit (A2) with a change of 12\% would cause three damage, a normal attack only one damage. For $\alpha=0.01$, the probabilities for a critical hit are irrelevant, as the balancer has equalized the damage for both cases. We discuss this in detail in combination with the influence of the parameter $\alpha$ in Section~\ref{sec:discussion}.

\section{Discussion and Limitations}
\label{sec:discussion}


The results of our experiments showed that the generation and value optimization for balancing of graph-based game economies with the proposed framework is feasible. There are, however, several points that need to be discussed.

The generator is controllable in terms of the number and types of nodes and generated valid graphs in terms of the given constraints, showing an average validity of 97\%. A median execution time of 25 ms indicates fast performance. In comparison to~\cite{rogers_using_2023}, our implementation is also able to construct game economy graphs that do not only represent tree structures and therefore allow loops or contain probabilistics, for instance. This allows for greater precision and flexibility in modeling economies~\cite{schreiber_game_2021,klint_micro-machinations_2013}. 
So far we only focused on the validity of generated graphs in relation to the node types. One approach for future work is therefore to focus on creating interesting or differently complex economies. 

We investigated on the general performance of the balancer by applying it to each economy in the generated dataset to a randomly chosen target value with a randomly chosen simulation length.
The results (Table~\ref{tab:gen-performance}) vary greatly dependent on the value of $\alpha$.
The best-balanced proportion yields $\alpha=0.05$ with a share of 93.3\%. $\alpha=0.0$ allows no margin for the simulated target values, thus its results are worse. Also, the execution time and needed number of generations differ per $\alpha$. Since the permissible margin of $\alpha=0.05$ allows for greater scatter, the target balance can be achieved faster and therefore its median generation and execution time are much faster compared to smaller $\alpha$.
However, there is no setting where all economy graphs could be balanced. In cases where the balancer improved overall, but could not achieve the expected quality, the algorithm may have gotten stuck in local optima. 

Another problem is the challenge of balancing randomly selected combinations of pools and values in combination with the distribution and networking of different node types. For instance, there are cases in which a certain target value cannot be mathematically achieved within the randomly chosen simulation length.
This could be addressed by interpreting the balancing constraint as a value range. In particular this is beneficial for use cases where the actual value is not important, but the perfect balance is.


With the case study of two fictional economies (mage and archer class), we delve into how the balancer optimizes both to ensure equal damage output in the same time frame, addressing the objective in game balancing for diverse strategies and preventing a single dominant strategy from consistently prevailing.
We compare the results of two runs with two different values of $\alpha$, representing the threshold for values to consider as balanced. For both configurations, the balancer could find a solution within a short number of generations. A key finding here is that for a low value of $\alpha$ (0.01), the algorithm tries to minimize the fluctuation of values caused through the stochastic simulation. While the probabilities for normal and critical hits still differ, it equalizes the damage for both hits and thereby mitigates the randomness. At the one hand, it fits the balancing criteria, at the other it might not be a desirable solution since now the intention of the economy design is obsolete. To address this shortcoming, we recommend using low values of alpha only if randomness should have no or little impact. In other words, $\alpha$ can be used not only to configure the precision to a specific value but also to modulate the stochastic impact.

Another solution for a game designer is to use static weights for e.g., one of the damage values to prevent the balancer from adjusting it.
For this case study we used the parametrization of the fitness function in Eq.~\ref{eq:f3} to balance two economies at once to a same value. In many cases, however, new game entities are to be integrated into an existing game ecosystem. Therefore, to not adjust the whole existing system, Eq.~\ref{eq:f1} can be used directly to balance the newly introduced content to a value which fits into the ecosystem.
Another point to mention is that, due to the recursive execution of the economy framework, the mage's economy implements the play style of a spammer, using an ability whenever its cooldown is ready and enough mana available. Human players would also use other strategies, such as waiting for an ability that deals more damage even though another one is available. A future approach is thus the implementation of small bots, each with different strategies for a same economy.

It was shown that the economy simulation framework is able to implement basic concepts of game economics on examples from the game Minecraft and two fictitious character classes. However, it lacks components to e.g., influence edge weights based on values of pools as implemented in~\cite{klint_micro-machinations_2013,vanrozen_adapting}. This would open further opportunities to study more complex economies and other balancing objectives such as the counteracting of positive feedback loops.

Another limitation is that this research is based on simulations on an abstract game. Since by design no real game is implemented and no humans for testing are involved, playtests are still required. It still depends on human players whether the generated economies with the computed weights are actually fun to play within the narrative setting chosen by the designer. Therefore, this research is intended to support the early stages of game design to find an interesting economy through generation and balancing initial values for the first player tests.
Lastly, evolutionary computing has a high dependency on randomness, especially in combination with probabilistics within the simulations. Hence, we attempt to mitigate this by evaluating GEEvo on a large sample of economies with a wide range of configuration values.

\section{Conclusion and Future Work}
We have proposed GEEvo (Game Economy Evolution), a framework for generating and balancing graph-based game economies in a two-step process using evolutionary algorithms and simulations.
In addition to a lightweight framework for simulating the economies for balancing, we presented a fitness function that can be parameterized differently to balance economies towards various objectives. The results show that the balancer can optimize the weights of the economies to arbitrary values and simulation lengths in most cases. We further evaluated GEEvo in a case study using fictional economies of two popular game character classes.
By considering game economies from an abstract perspective, GEEvo is independent of a specific game or genre and is intended to support designers in the early stages of development.

In future work, we aim to extend the balancer to be able to handle multi-object evolution to meet multiple balancing goals for an economy simultaneously and expand its usage with further fitness functions. In addition, we want to extend the simulation framework to handle more complex economies, e.g., to influence an edge's weight depending on the value of a pool. This would open new opportunities to investigate how positive feedback loops could be counteracted by the balancer. Furthermore, an investigation into the automated naming of an economy's nodes based on a given narrative context is of interest through recent advances in large language models (LLMs).

\label{sec:conclusion}

\bibliographystyle{unsrt}
\bibliography{cog.bib}

\end{document}